\documentclass[10pt,twocolumn,letterpaper]{article}

\usepackage{wacv}              % To produce the CAMERA-READY version
\usepackage{graphicx}
\usepackage{amsmath}
\usepackage{amssymb}
\usepackage{booktabs}
\usepackage{xcolor}
\usepackage{multirow}

\usepackage[pagebackref,breaklinks,colorlinks]{hyperref}

\usepackage[capitalize]{cleveref}
\crefname{section}{Sec.}{Secs.}
\Crefname{section}{Section}{Sections}
\Crefname{table}{Table}{Tables}
\crefname{table}{Tab.}{Tabs.}

\def\wacvPaperID{1341} % *** Enter the WACV Paper ID here
\def\confName{WACV}
\def\confYear{2024}

\begin{document}

\title{ENTED: Enhanced Neural Texture Extraction and Distribution for Reference-based Blind Face Restoration}

\author{Yuen-Fui Lau\\
HKUST\\
{\tt\small yflauad@connect.ust.hk}
\and
Tianjia Zhang\\
HKUST\\
{\tt\small tzhangbl@connect.ust.hk}
\and
Zhefan Rao\\
HKUST\\
{\tt\small zraoac@connect.ust.hk}
\and
Qifeng Chen\\
HKUST\\
{\tt\small cqf@ust.hk}
}

\maketitle

%%%%%%%%% ABSTRACT
\begin{abstract}
We present ENTED, a new framework for blind face restoration that aims to restore high-quality and realistic portrait images. Our method involves repairing a single degraded input image using a high-quality reference image. We utilize a texture extraction and distribution framework to transfer high-quality texture features between the degraded input and reference image. However, the StyleGAN-like architecture in our framework requires high-quality latent codes to generate realistic images. The latent code extracted from the degraded input image often contains corrupted features, making it difficult to align the semantic information from the input with the high-quality textures from the reference. To overcome this challenge, we employ two special techniques. The first technique, inspired by vector quantization, replaces corrupted semantic features with high-quality code words. The second technique generates style codes that carry photorealistic texture information from a more informative latent space developed using the high-quality features in the reference image's manifold. Extensive experiments conducted on synthetic and real-world datasets demonstrate that our method produces results with more realistic contextual details and outperforms state-of-the-art methods. A thorough ablation study confirms the effectiveness of each proposed module.
\end{abstract}

%%%%%%%%% BODY TEXT
\section{Introduction}
\label{sec:intro}
\begin{figure*}[h!]
\centering
\begin{tabular}{c@{\hspace{0.1mm}}c@{\hspace{0.1mm}}c@{\hspace{0.1mm}}c@{\hspace{1.0mm}}c@{\hspace{0.1mm}}c@{\hspace{0.1mm}}c@{\hspace{0.1mm}}c@{\hspace{0.1mm}}c@{\hspace{0.1mm}}}
\includegraphics[width=0.125\linewidth]{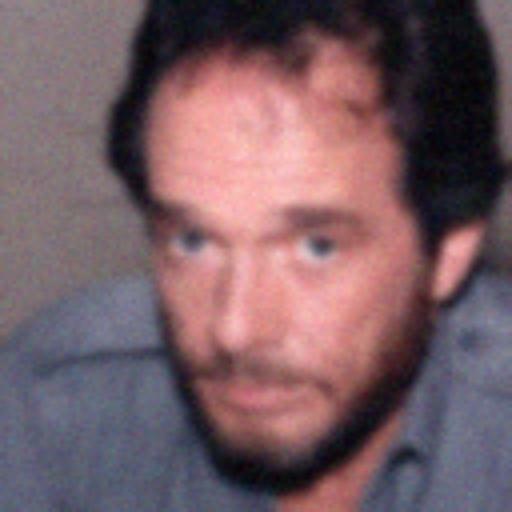}&
\includegraphics[width=0.125\linewidth]{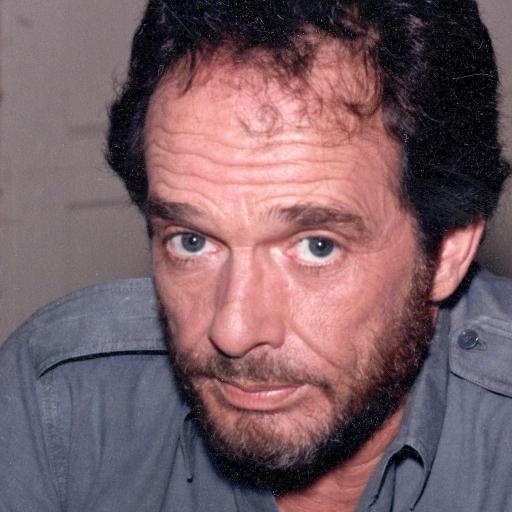}&
\includegraphics[width=0.125\linewidth]{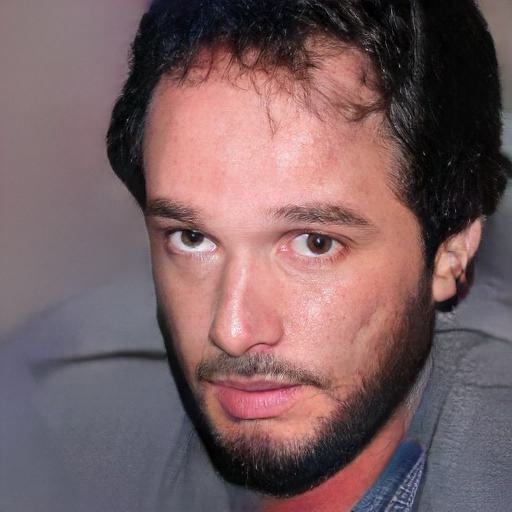}&
\includegraphics[width=0.125\linewidth]{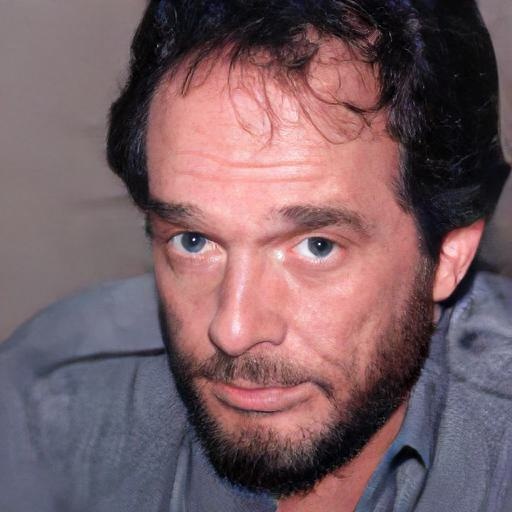}&
\includegraphics[width=0.125\linewidth]{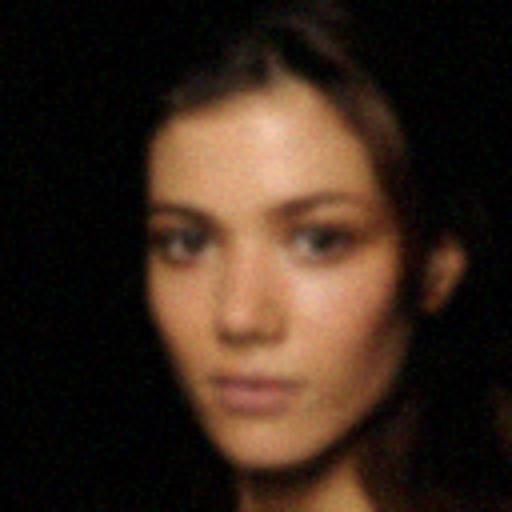}&
\includegraphics[width=0.125\linewidth]{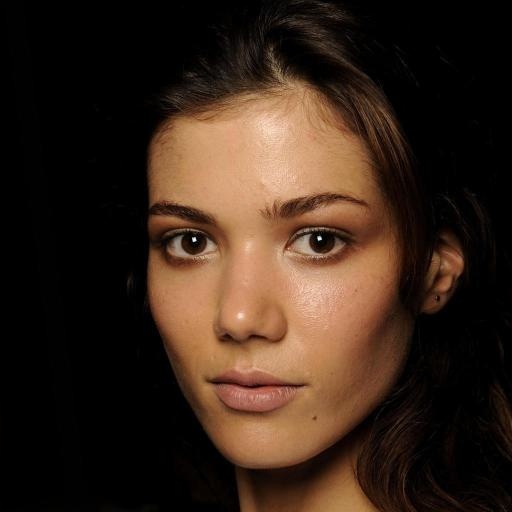}&
\includegraphics[width=0.125\linewidth]{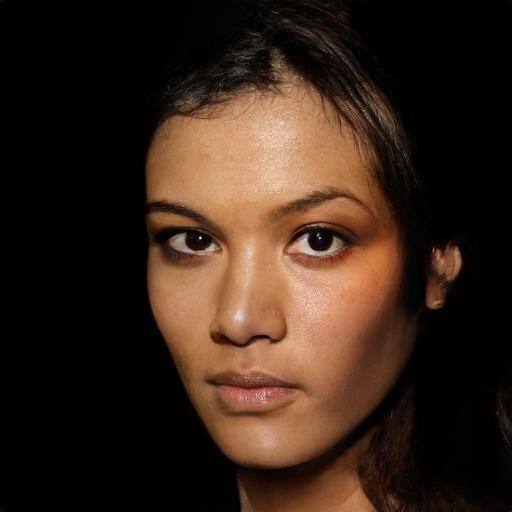}&
\includegraphics[width=0.125\linewidth]{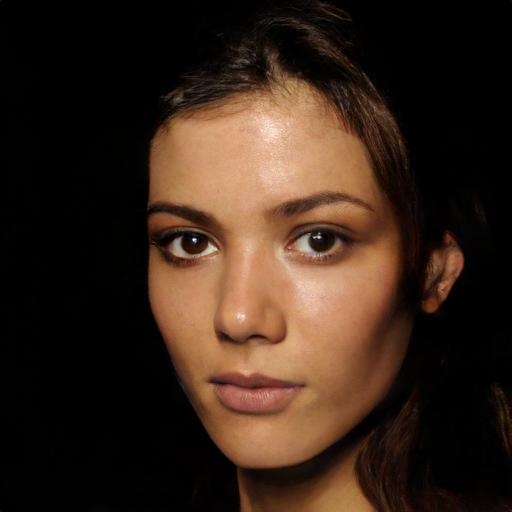}&
\\
\includegraphics[width=0.125\linewidth]{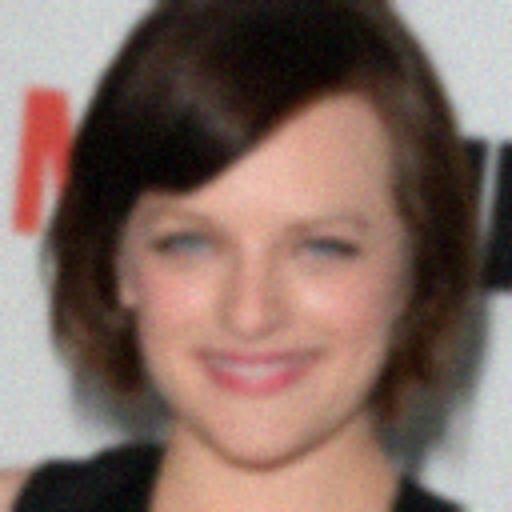}&
\includegraphics[width=0.125\linewidth]{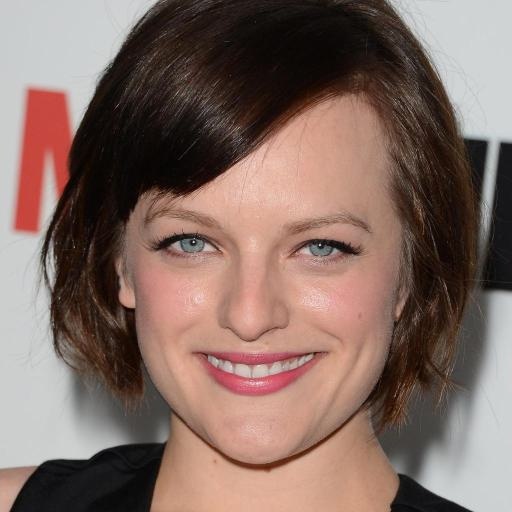}&
\includegraphics[width=0.125\linewidth]{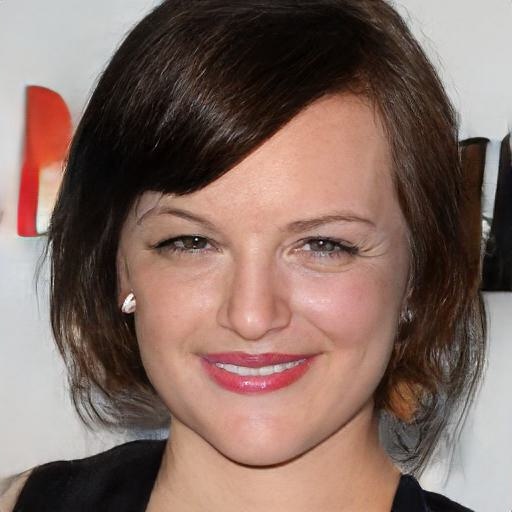}&
\includegraphics[width=0.125\linewidth]{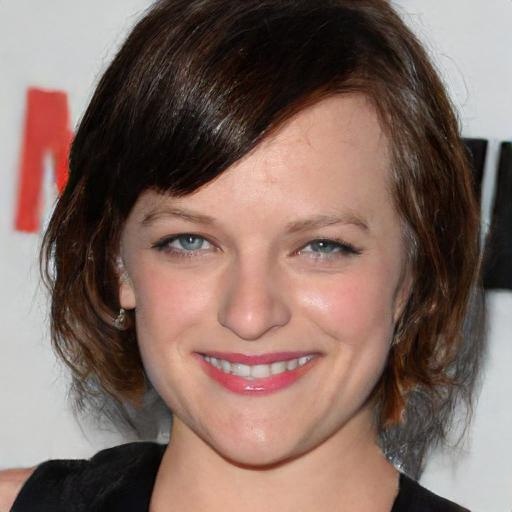}&
\includegraphics[width=0.125\linewidth]{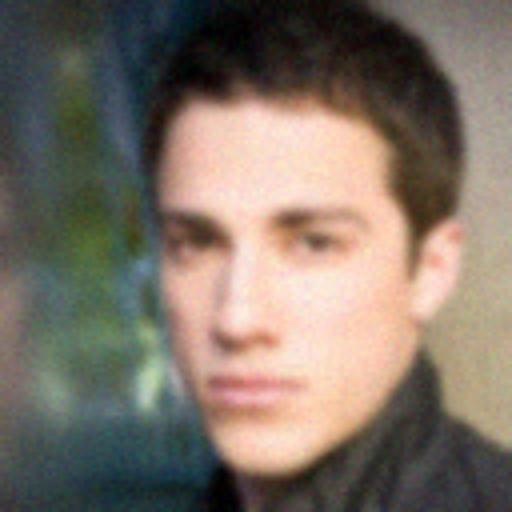}&
\includegraphics[width=0.125\linewidth]{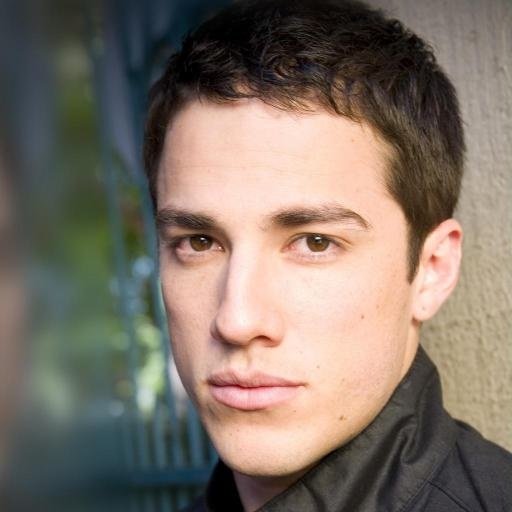}&
\includegraphics[width=0.125\linewidth]{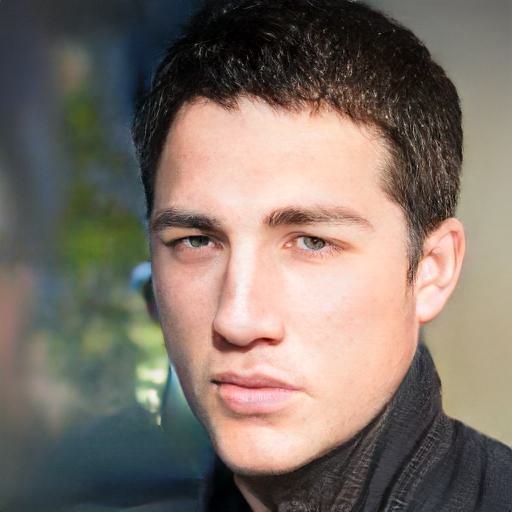}&
\includegraphics[width=0.125\linewidth]{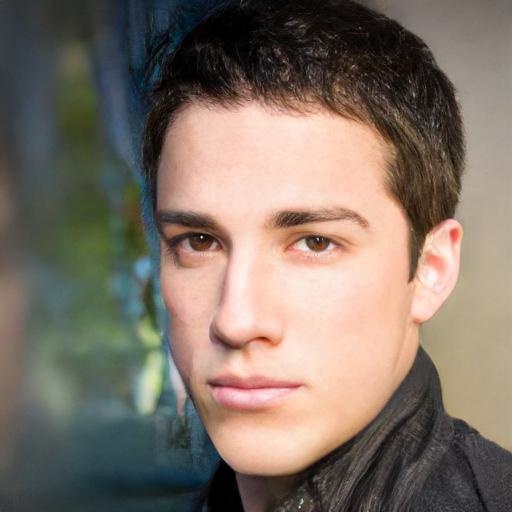}&
\\
Input & Ground truth & Without residual & With residual  &  Input &  Ground truth & Without residual  & With residual \\
%image& & connection & connection &image & & connection & connection

\end{tabular}

\caption{When each face restoration is performed using the same reference image and experimental setup, the output without a residual connection tends to degrade the facial identity. However, when a residual connection is used, the result exhibits superior facial details that align more closely with the original image.}
\label{fig: skip_and_no_skip}

\end{figure*}
Blind face restoration (BFR) is a technique in computational photography that focuses on transforming low-quality facial images into high-quality ones, even when the type of degradation is unknown. This process is crucial for those who wish to improve the quality of their facial images. However, the task of accurately reconstructing facial details from a low-quality image can be quite challenging due to the loss of specific identity information. To tackle this issue, we consider the use of a high-quality reference image of the same individual. This approach leads us to explore the concept of reference-based blind face restoration, where our goal is to utilize the information from a high-quality reference image to enhance the restoration process.
% Blind face restoration (BFR) is a computational photography application that involves restoring high-quality (HQ) face images from low-quality ones with unknown degradation. This task is important for people who want to enhance the quality of their face images. However, when given a low-quality (LQ) face image, it becomes challenging to reconstruct the facial details accurately due to the loss of identity-specific information. To address this challenge, we explore the possibility of using a high-quality reference face image of the same person. This leads us to investigate the problem of reference-based blind face restoration, where we aim to leverage the information from a high-quality reference image to improve the restoration process.

%Due to the complexity and unpredictability of the degradation in real-world scenarios, such as motion blur, noise, low resolutions, the restoration work becomes challenging.  
The technique of reference-based super-resolution (RefSR) has been gaining interest recently. It enhances the quality of a low-resolution input by incorporating high-quality semantic details from a reference image. However, if the features of the reference image are not managed correctly, it can result in under-utilization or incorrect use of the reference. To overcome these challenges, we employ a texture extraction and distribution framework that can be trained with attention reconstruction loss to improve the accuracy and the use of HQ features throughout the texture transfer process. The texture extraction and distribution framework was successfully applied to the controllable human image synthesis task \cite{ren2022neural}, and we extend this concept to our reference-based blind face restoration framework. 
 
The latent representation of low-quality (LQ) input often contains incorrect information during the texture distribution process. Simply applying the texture extraction and distribution framework is not enough to produce high-fidelity images. To overcome this challenge, we utilize a vector-quantization (VQ) technique. This technique entails replacing degraded latent features with high-quality latent codes obtained from a high-quality (HQ) dictionary. By directly substituting these codes, we are able to close the disparity between low and high-quality latent codes, thereby offering suitable semantic guidance for texture distribution.

As shown in table \ref{T:ablation}, we notice that modulated convolutions aid in enhancing the realism of facial details during the restoration process. However, they necessitate high-quality style code representation for superior image restoration. Generating meaningful style codes from a high-dimensional latent space is challenging \cite{tov2021designing}, particularly with an imperfect and noisy latent representation of the degraded input image. To tackle this, we refine the latent space using the cross-attention technique \cite{chu2022nafssr} and information from the reference prior. This cross-comparison and refinement of the low-quality latent features with the HQ latent codes from the reference prior yield an enriched latent space. This assists in creating HQ style codes with more significant texture detail. Different from previous works \cite{richardson2021encoding, menon2020pulse} for blind face restoration, which typically infers the style code from a single information source, we utilize a reference-based method, using the HQ features from the reference image to refine the latent space. This results in a higher quality style code generation with fewer flawed semantic features and biases. 

We also recognize that a degraded input image loses not just texture information but essential structural content information, crucial for maintaining fidelity. As demonstrated in figure \ref{fig: skip_and_no_skip}, insufficient fidelity information or incomplete inference for fidelity features leads to a significant loss of facial identity. To counteract this, we add residual connections to enhance the flow of fidelity information between the decoder and content encoder, thereby preserving identity.

Our contribution can be summarized as follows:
\begin{itemize}
    \vspace{-1mm}
    \item We apply a latent space refinement technique that can utilize the information from the high-quality reference prior for generating style code that carries high-quality semantic details.
    \item To facilitate the texture distribution process, we modify the texture extraction and distribution framework by adopting the vector-quantization technique.
    \vspace{-1mm}
    \item Extensive experiments show that our model advances the state-of-the-art method in blind face restoration. It can handle badly corrupted face images from both real-world and synthetic data.
\end{itemize}

\section{Related Work}
\begin{figure*}[h!]
\centering
\begin{tabular}
{c@{\hspace{0.1mm}}c@{\hspace{0.1mm}}c@{\hspace{0.1mm}}c@{\hspace{0.1mm}}c@{\hspace{0.1mm}}c@{\hspace{0.1mm}}c@{\hspace{0.1mm}}c@{\hspace{0.1mm}}}
\includegraphics[width=0.135\linewidth]{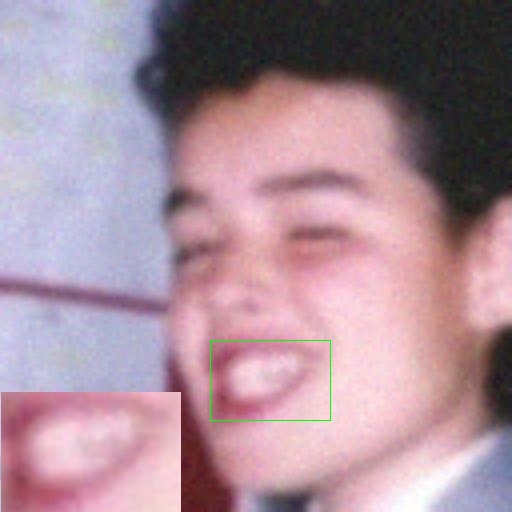}&
\includegraphics[width=0.135\linewidth]{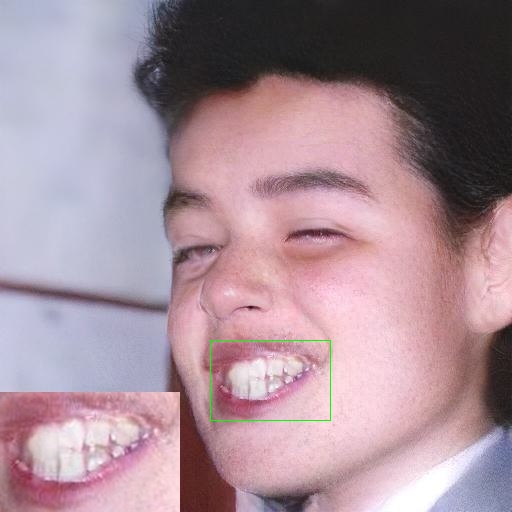}&
\includegraphics[width=0.135\linewidth]{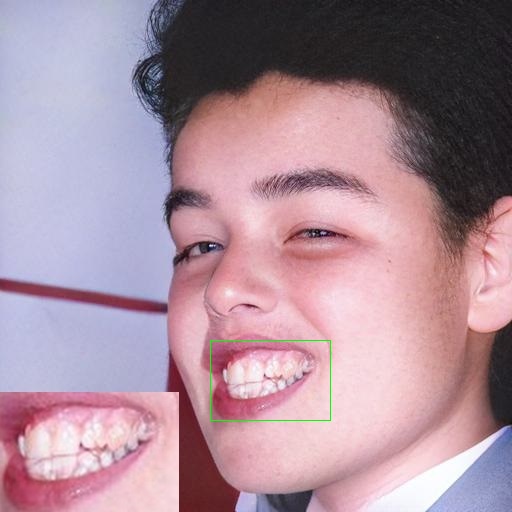}&
\includegraphics[width=0.135\linewidth]{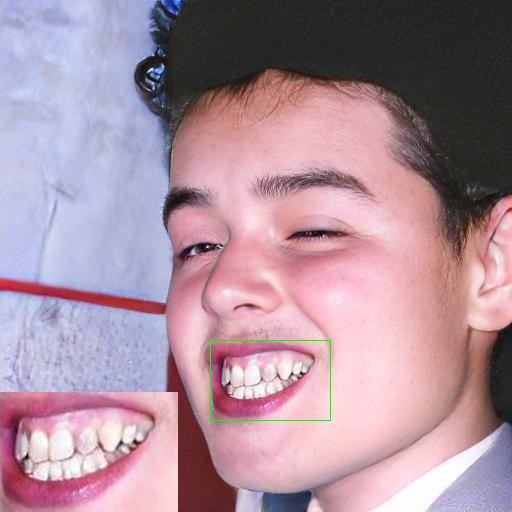}&
\includegraphics[width=0.135\linewidth]{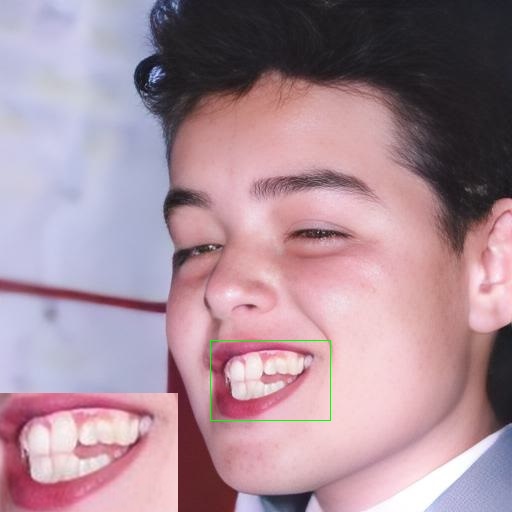}&
\includegraphics[width=0.135\linewidth]{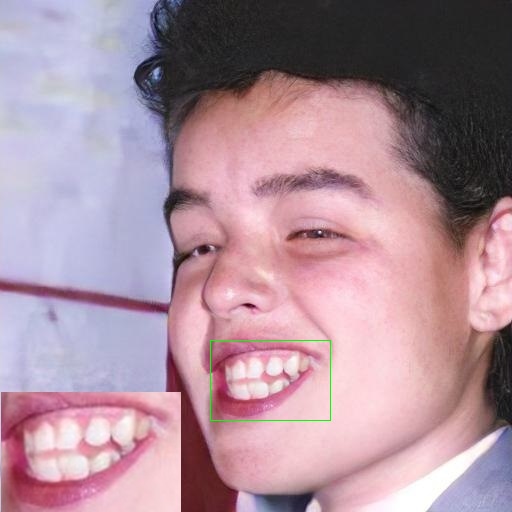}&
\includegraphics[width=0.135\linewidth]{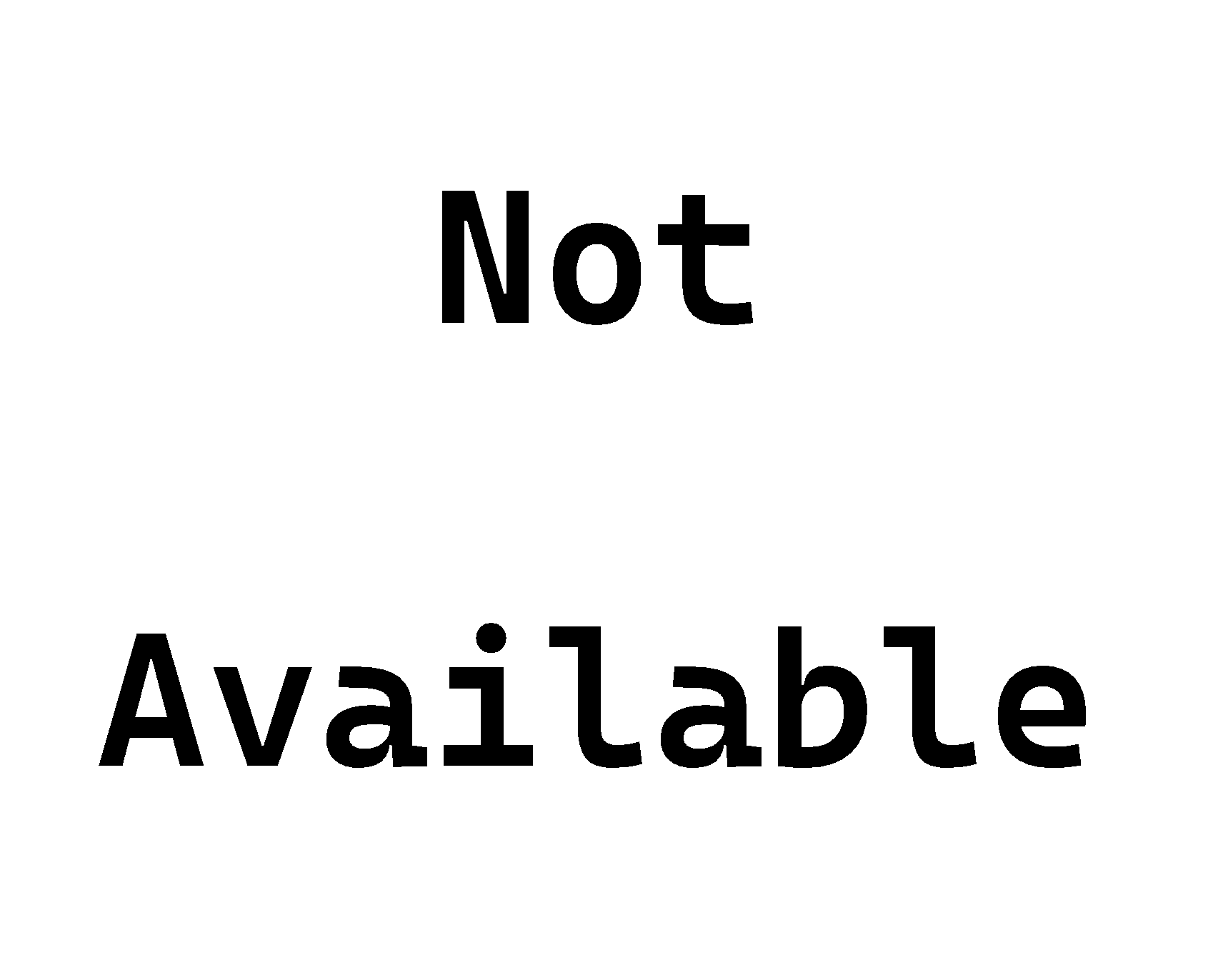}&
\\
\includegraphics[width=0.135\linewidth]{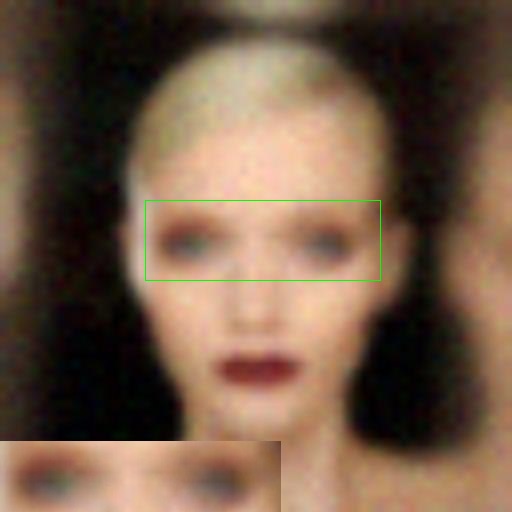}&
\includegraphics[width=0.135\linewidth]{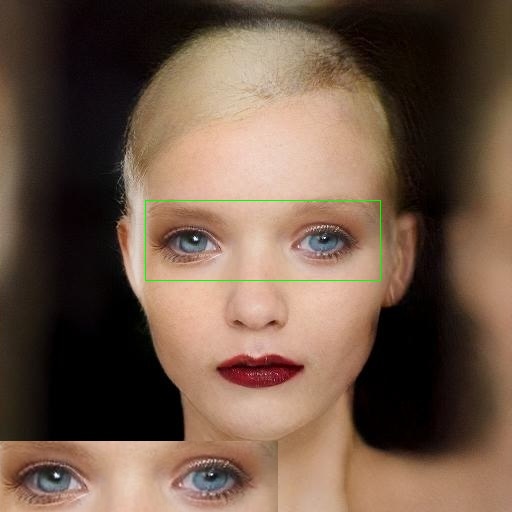}&
\includegraphics[width=0.135\linewidth]{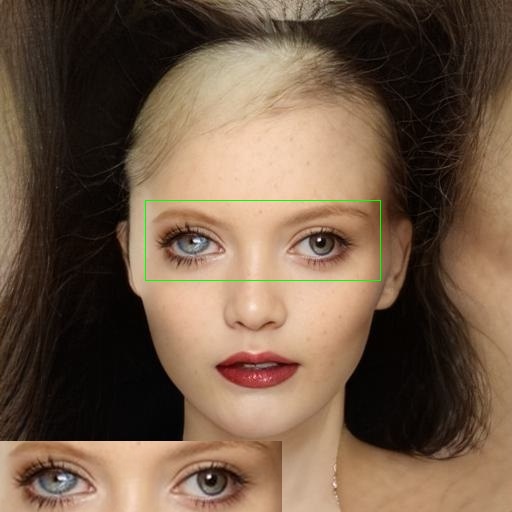}&
\includegraphics[width=0.135\linewidth]{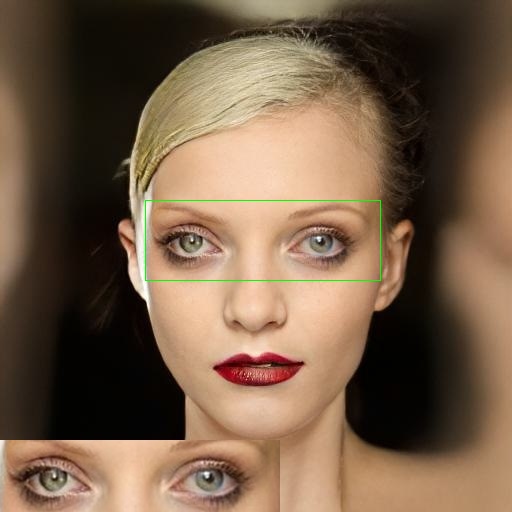}&
\includegraphics[width=0.135\linewidth]{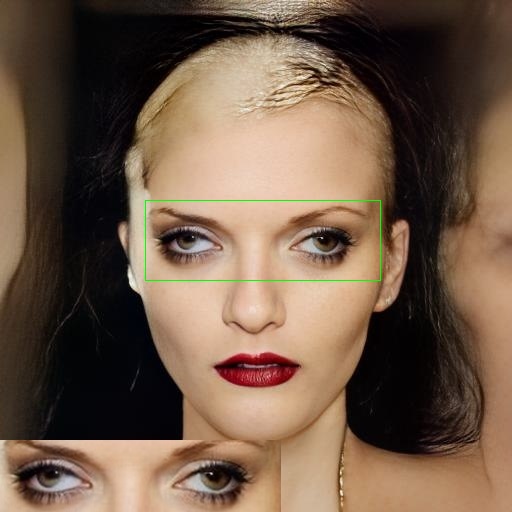}&
\includegraphics[width=0.135\linewidth]{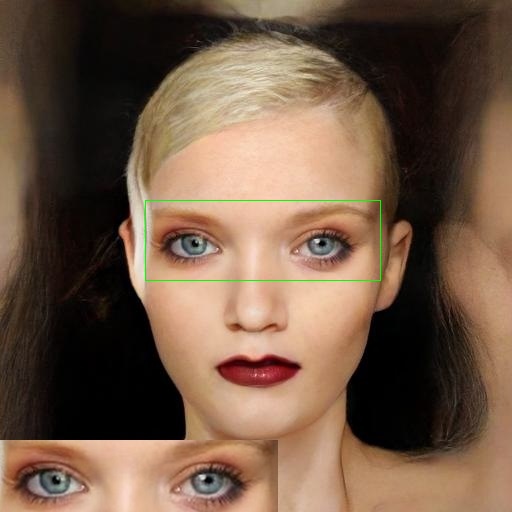}&
\includegraphics[width=0.135\linewidth]{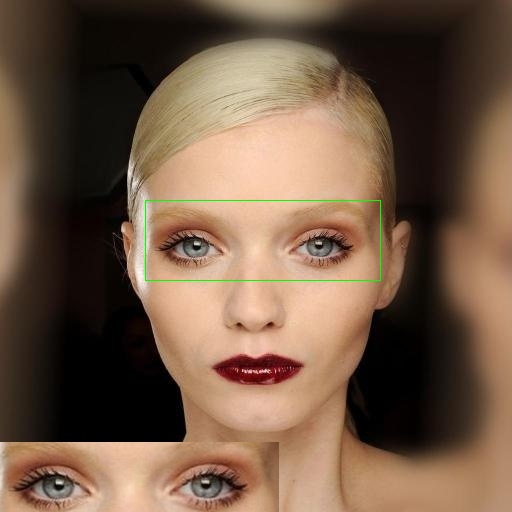}&
\\
Input & DFDNet & GPEN & VQFR & GFP-GAN & \textbf{ENTED} &Ground truth
% & &  &  & & &  & 
\end{tabular}
\caption{Visual comparison with the state-of-the-art blind face restoration methods. The first row demonstrates the visual comparisons of real-world data and the second row shows the visual comparisons of synthesized data.}
\end{figure*}

\subsection{Blind Face Restoration}
The goal of Blind Face Restoration (BFR) is to restore high-quality faces that have suffered from complicated and unknown degeneration. In recent years, it has gotten a lot of attention. Since Human perceptions are more sensitive to facial images than other image domains, hence necessitating more solid and thorough image generation manipulation. Most high-resolution face restoration methods \cite{wang2022restoreformer, richardson2021encoding, gu2022vqfr} use maximum likelihood estimation to rebuild realistic facial features and adversarial learning to produce a natural image distribution. Prior face restoration methods often make use of face-specific priors, such as generative priors \cite{gu2020image,menon2020pulse,wan2020bringing, wang2021towards, richardson2021encoding}. These techniques typically use a pre-trained face generative adversarial network such as StyleGAN \cite{karras2019style, karras2020analyzing} and embed the latent representation of the degraded image into the GAN latent space, which encapsulates rich and diverse high-quality features like facial texture and facial geometry, etc. Some of them are done by projecting the degraded face straight into the latent space \cite{wan2020bringing, wang2021towards} or exploring a latent vector with a costly target-specific optimization \cite{menon2020pulse}. Directly embedding the latent representation of a degraded input image into the latent space of a pre-trained GAN model, however, tends to generate overly smooth face images and inaccurate facial details because the latent representation of a degraded input image carries noise and leads to the formation of imperfect latent codes with erroneous information. Another approach explores geometric priors, for which facial landmarks \cite{chen2018fsrnet, kim2019progressive}, face parsing maps \cite{chen2021progressive, shen2018deep}, and facial component heat-maps \cite{yu2018face} are commonly employed in blind face restoration algorithms. Because the bulk of the geometric information is approximated from distorted faces, and the geometric information becomes imprecise for substantially degraded input images, the restoration performance is easily influenced by the input's quality. It does not function well for generating realistic facial details in extreme cases.

\subsection{Reference-based Restoration}
As corrupted images often lack high-quality details, the reference-based image restoration approach seeks to extract high-quality features from the reference image to be transferred to the low-quality image. Unlike Single Image Super-Resolution (SISR), which does not include extra information, the Reference-based Image Super-Resolution (RefSR) task restores damaged input images by transferring high-quality features from reference images. In SRNTT \cite{zhang2019image}, the correspondence matching was done based on the extracted features from the pre-trained VGG network. Xie et al. ~\cite{yang2020learning} and Yang et al. ~\cite{xie2020feature} employed learnable feature extractors that undergo end-to-end training with the primary restoration network. However, their correspondences were calculated only using contents and appearances. Recent works like Dogan et al. ~\cite{dogan2019exemplar} employ a warper sub-network to align the content of the input and reference images in order to perform feature fusion. Li et al. ~\cite{Learning_Warped} proposed an integrated framework for face restoration, which includes a warping subnetwork to predict flow fields for transferring features and an extra reconstruction subnetwork to utilize the warped features and then recover the clear images. Subsequently, Li et al.~\cite{EBFR_Multi_Exemplar} increase the restoration quality by using more reference face images and an adaptive mechanism to fuse the high-quality guidance features and low-quality features. Recently Li et al. \cite{Dual_Memory} turn the high-quality features into two dictionaries encoding generic high-quality prior information and person-specific features, respectively. Texture mapping methods \cite{lu2021masa, yang2020learning} maps the high-quality texture from the reference image to the corrupted input image. In order to align the features of the input and reference images, Yang et al.  ~\cite{yang2020learning} and Zhang et al. ~\cite{zhang2019image} adopt the Patch-Matching technique. However, there are transformation gaps and resolution gaps between the input and reference images. \cite{jiang2021robust} performs correspondence matching in a different way. It employs a contrastive correspondence network and knowledge distillation to bridge the gap. 

\begin{figure*}[h!]
\centering
  \includegraphics[width=15cm]{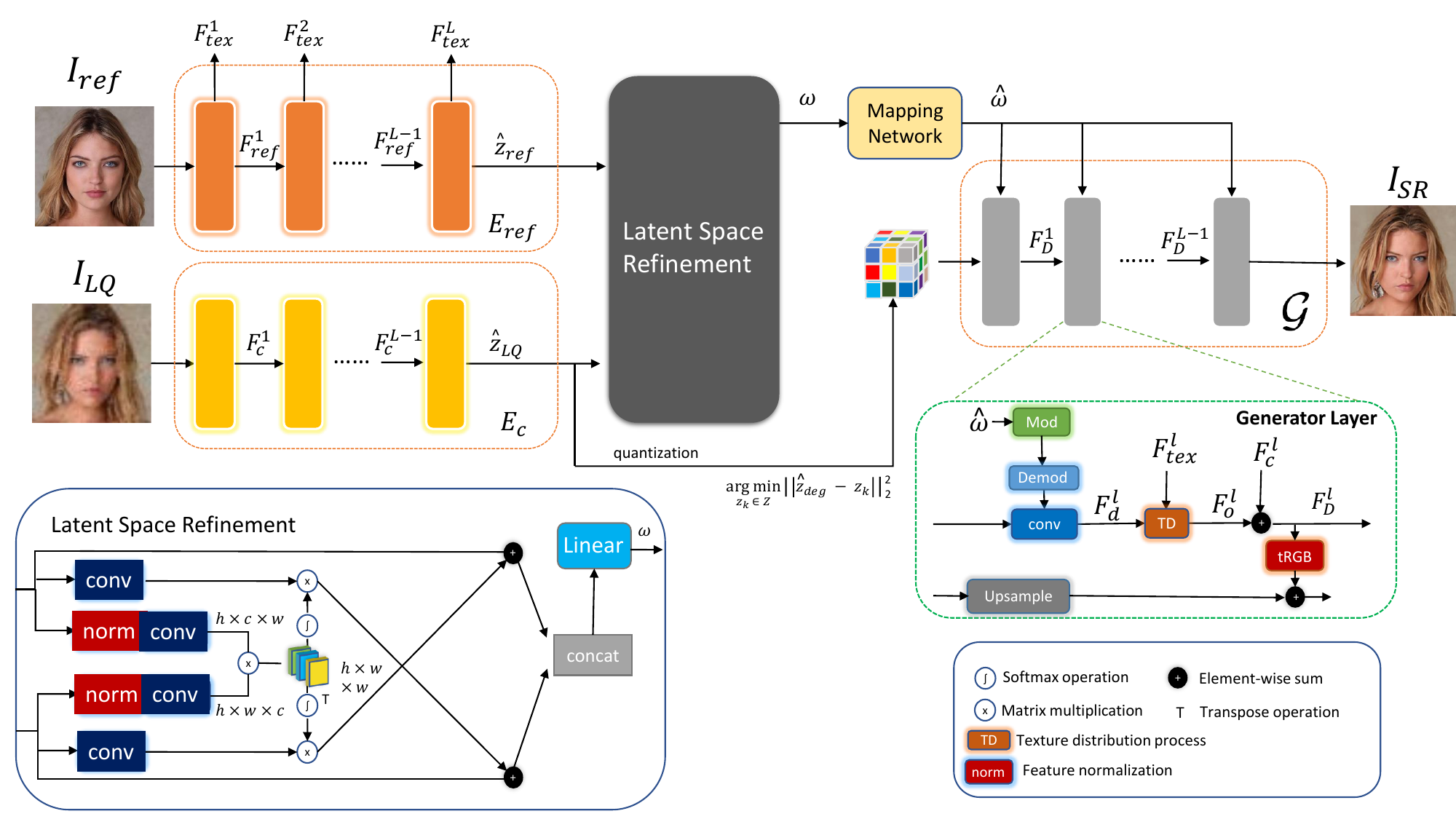}
  \caption{A summary of our pipeline. Using reference features, we construct a high-quality image by transferring high-quality reference details (Neural Texture Extraction and Distribution Process) and repairing (Application of VQ Dictionary and Latent Space Refinement) distorted semantic information in degraded input images.}
\end{figure*}

\section{Method}
\subsection{Neural Texture Extraction and Distribution}
\label{sec3_1}

One of the most difficult aspects of texture transfer is precisely reassembling the texture extracted from the reference images and properly mapping them to the degraded input images. To address this issue, neural texture extraction and distribution ~\cite{ren2022neural} based on the attention mechanism is introduced.  The benefit of using ~\cite{ren2022neural} for texture transfer is that it helps to learn specific extraction and distribution kernels to map appropriate semantic texture components between the input and reference images. During the extraction process, high-quality semantic textures extracted from the reference image act as keys and values, while the features obtained from the low-quality input image work as queries, providing clues that guide appropriate high-quality semantic textures to enhance the corrupted textures in the input image.

\textbf{Texture Extraction Process.} 
Firstly, a texture encoder $E_{ref}$ is employed to extract a hierarchy level of feature maps from the reference image $I_{ref}$,  denoted as $F_{ref}^{i}$, where $i$ represents the layer index. Then we utilize a set of texture extraction kernels $W^{i}_e$ to project the feature maps into latent texture codes, $C^{i}_{e} = W^{i}_e F^i_{ref}$, where $W^{i}_e \in \mathbb{R}^{k \times c}$ and $F_{ref}^{i}\in \mathbb{R}^{c \times hw}$. Here we compress the height and width dimensions into one spatial dimension to better apply matrix multiplication. Softmax is applied on $C^{i}_e$ along the spatial dimension to obtain semantic texture extraction matrix $\tilde{C}_e^{i}$. The neural texture $F_{tex}^i \in \mathbb{R}^{k \times c}$ is obtained as:  
\begin{equation}
\begin{aligned}
\tilde{C}_{e}^{i}(p,q) &= \frac{exp(C^{i}_{e}(p,q))}{\sum^{hw}_{q=1} exp(C^{i}_{e}(p,q))},\\
F_{tex}^{i} &= (\tilde{C}^{i}_e) ( f_{1\times1}({F_{ref}^{i}}) )^{\intercal},
\end{aligned}
\end{equation}
\noindent 
where $f_{1\times 1}(\cdot)$ indicates a $1\times 1$ convolution, p and q are the indices along the channel and spatial dimensions, respectively.

\textbf{Texture Distribution Process.} To distribute the extracted semantic texture to the low-quality input image $I_{LQ} \in \mathbb{R}^{3 \times H \times W}$, a content encoder  $E_{c}$  is first applied to $I_{LQ}$ to get the high-level semantic feature map of the input image. The final output of $E_{c}$ is then directly passed into a decoder G. Let $F_{d}^{i}$ denote the output feature map from the $i^{th}$ layer of the decoder G. The distribution process starts in the same manner as the extraction operation. The latent codes $C_{d}^{i} = W_{d}^{i}F_{d}^{i}$ is obtained by applying texture distribution kernels $W_{d}^{i} \in \mathbb{R}^{k \times c}$ on $F_{d}^{i}\in \mathbb{R}^{c \times hw}$. The resulting feature map$ F_{o}^{i} \in \mathbb{R}^{c \times hw}$  is obtained by first adopting softmax function on $C_{d}^{i}$ along the channel dimension and then multiplying the transpose of extracted neural texture $F_{tex}$ with the semantic texture distribution matrix $\tilde{C}_{d}^{i}$. The overall distribution process is formulated as follows:
\begin{equation}
\begin{aligned}
\tilde{C}_{d}^{i}(p,q) &= \frac{exp(C^{i}_{d}(p,q))}{\sum^{k}_{p=1} exp(C^{i}_{d}(p,q))},\\
F_{o}^{i} &=(F_{tex}^{i})^{\intercal} (\tilde{C}_{d}^{i}) .
\end{aligned}
\end{equation}

\subsection{Vector-Quantized Dictionary}
\label{VQD}
Some prior works \cite{gu2022vqfr, wang2022restoreformer, li2020blind} have introduced a vector quantized dictionary to replace faulty features in the latent space for blind face restoration. They consider the dictionary a high-quality low-level feature bank that stores meaningful facial representations. Different from prior work \cite{gu2022vqfr, wang2022restoreformer, zhou2022towards, chen2022blind, zhao2022rethinking} that consider the dictionary as a feature bank for facial representation, we treat the HQ dictionary as a reservoir of semantic queries that can provide comprehensive guidance during the texture distribution process. We define the HQ dictionary with $K$ discrete tokens as $\mathcal{Z} = \{z_{k}\}^{K}_{k=1}$ . The feature map of the last layer obtained from the content encoder $Z_{LQ} = E_{c}(I_{LQ}) \in \mathbb{R}^{c \times h \times w}$  could be considered as a set of vectors of length cc expanded along the spatial dimension, denoted by $\{z_{LQ}^{i,j}\}_{i=1,j=1}^{h,w}$. Subsequently, each vector can be enhanced by replacing it with the nearest entry $\hat{z}_{k} \in \mathbb{R}^{c}$ in the HQ dictionary $\mathcal{Z}$, which has the minimum $l_{2}$ distance with the feature vector $\hat{z}^{i,j}_{LQ} \in \mathbb{R}^{c}$ located at spatial position (i, j), The overall quantization process $q(\cdot)$ is defined as    
\begin{equation}
z_{q}^{i,j} = q(z^{i,j}_{LQ}, \hat{z}_k) = \mathop{\arg\min}_{z_k \in \mathcal{Z}}  \| z^{i,j}_{LQ} - z_{k} \|^{2}_{2},
\end{equation}
\noindent where $z_{q}^{ij} \in \mathbb{R}^{c}$ is the quantized feature vector, and all the quantized vectors could be rearranged back as a refined feature map $\hat{Z}_{LQ}$.

Since the quantization process is non-differentiable, we follow a similar optimization strategy as VQ-GAN \cite{esser2021taming}. A straight-through gradient estimator is employed to  simply copies the gradients from the decoder to the content encoder and optimize via the quantization loss $\mathcal{L}_{q}$:
\begin{equation}
\mathcal{L}_{q} = \sum^{h}_{i=1} \sum^{w}_{j=1} \|  sg[z^{i,j}_{LQ}] - z_{q}^{i,j}  \|^{2}_{2} + \beta \|  sg[z_{q}^{i,j}] - z^{i,j}_{LQ}  \|^{2}_{2},
\end{equation}
where $\beta$ is the commitment weight and $sg[\cdot]$ denotes the stop gradient operator. Throughout all the experiments, the commitment weight is set to 0.25.

Given a degraded input image  $I_{LQ} \in \mathbb{R}^{3 \times H \times W}$, the degraded latent representation $\hat{z}_{LQ} = E_{c}(I_{LQ}) \in \mathbb{R}^{c \times h \times w}$ , where $h = H/s$, $w = W/s$, and s is the down-sampling factor. Following previous works \cite{gu2022vqfr, esser2021taming}, we choose $s = 32$ for $512 \times 512$ input. Moreover, in order to maximize the usage of the HQ dictionary, we adopt~\cite{lancucki2020robust} and periodically re-initialize the dictionary with the k-means algorithm as what it did in ~\cite{gu2022vqfr}. The number of clusters $k$ is chosen to be 1024. 

%tor. We also find that reconstruction quality, as mentioned in previous work \cite{gu2022vqfr, esser2021taming}, is influenced by the down-sampling factor ss. To achieve a better image quality in terms of facial texture,

\subsection{Latent Space Refinement}
\label{AttLE}
Our method applies style code and modulated convolution proposed in the StyleGAN\cite{karras2019style} architecture for restoration. However, generating the style code $\omega \in \mathcal{W} \subseteq \mathbb{R}^{c}$  using only high-quality features from the reference image leads to the loss of fidelity information, which is essential for retaining facial identity. Nonetheless, generating the style code using only latent features from degraded input introduces erroneous information, which is harmful to the perceptual quality of the final restored image. To minimize the fidelity change while maintaining high perceptual quality, we generate the style code $\omega$ from the refined latent space of $\hat{z}_{LQ}$ and $\hat{z}_{ref}$, which is achieved by exchanging information between them through a cross attention technique. We follow a similar formulation as~\cite{chu2022nafssr} with a bit of modification on the normalization method, which results in more stable training. We define the scalar dot product attention \cite{vaswani2017attention} as $\phi$ with given query $Q$, key $K$ and value $V$:  
%This is achieved by replacing corrupted features in \hat{z}_{deg}\hat{z}_{deg} with the high-quality features derived from the reference image's manifold while 
\vspace{-2mm}
\begin{equation}
\phi_{\lambda}(Q, K, V) = Softmax(\frac{Q K^{\intercal}}{\sqrt{d}}) V,
\end{equation}
where $\lambda$ indicates applying the Softmax function along $\lambda$ axis and  $d$ is a normalization constant.
Given an input image $I_{LQ} \in \mathbb{R}^{3 \times H \times W}$ and a reference image  
$I_{ref} \in \mathbb{R}^{3 \times H \times W}$, 
we pass them into the content encoder $E_{c}$ and the texture encoder $E_{ref}$ separately to get the semantic latent representation 
$\hat{Z}_{LQ} = E_{c}(I_{LQ}) \in \mathbb{R}^{c \times h \times w}$ and $\hat{Z}_{ref} = E_{ref}(I_{ref}) \in \mathbb{R}^{ c \times h \times w }$. 
Then, we apply feature normalization \cite{ren2022neural} $FN$ on $\hat{z}_{LQ}$ and $\hat{z}_{ref}$ for stable training. The feature normalization is formulated as follows:
\begin{equation}
FN_{\Lambda}(z) = \frac{x - \mathbb{E}_{\Lambda}(x)}{\sqrt{\sum_{i=1}^{\Lambda} x^{2}_{i} + \epsilon }},
\end{equation}
where $\Lambda$ indicates the operation along the $\Lambda$ axis (e.g., the channel axis), $\epsilon$ is a constant to avoid numeric issues, and $\mathbb{E}_{\Lambda}(x)$ represents the expectation value of $x$ along the $\Lambda$ axis. After that, the quantized feature maps are processed by 1x1 convolution and reshaped along the spatial dimension to get the feature matrices $\tilde{Z}_{LQ}$, $\tilde{Z}_{ref}\in \mathbb{R}^{c\times hw}$ for attention calculation:
\begin{figure*}[h!]
\centering
    \begin{tabular}{c@{\hspace{0.1mm}}c@{\hspace{0.1mm}}c@{\hspace{0.1mm}}c@{\hspace{0.1mm}}c@{\hspace{0.1mm}}c@{\hspace{0.1mm}}c@{\hspace{0.1mm}}}
    \includegraphics[width=0.142\linewidth]{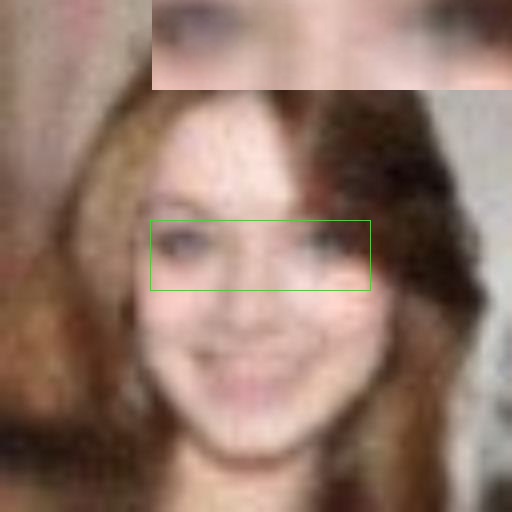}&
    \includegraphics[width=0.142\linewidth]{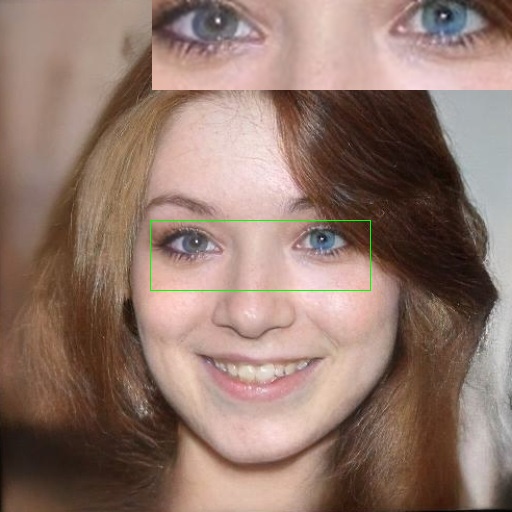}&
    \includegraphics[width=0.142\linewidth]{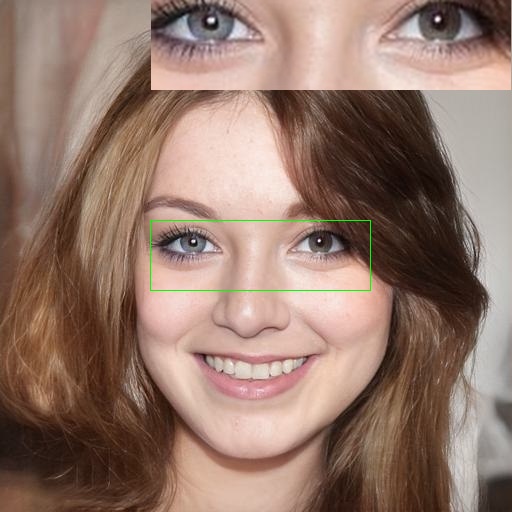}&
    \includegraphics[width=0.142\linewidth]{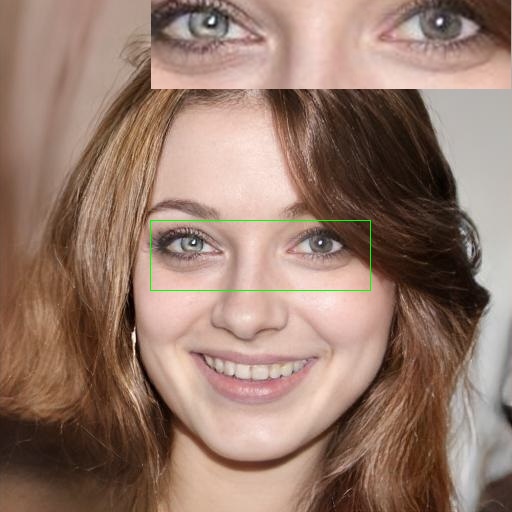}&
    \includegraphics[width=0.142\linewidth]{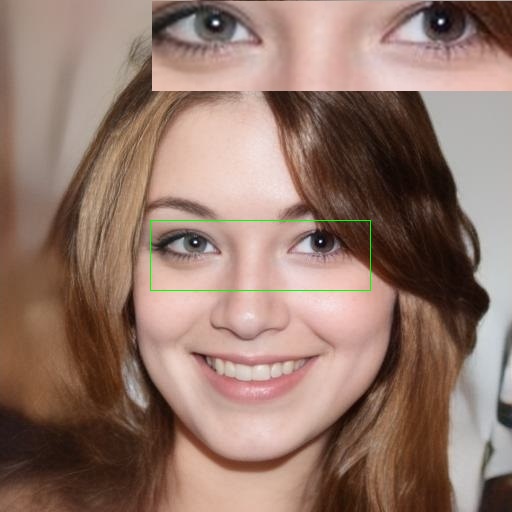}&
    \includegraphics[width=0.142\linewidth]{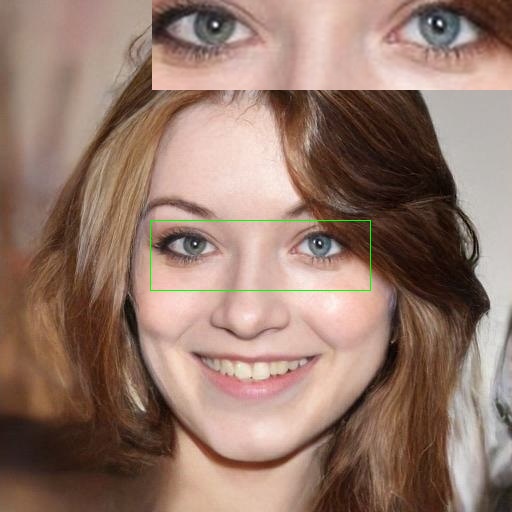}&
    \includegraphics[width=0.142\linewidth]{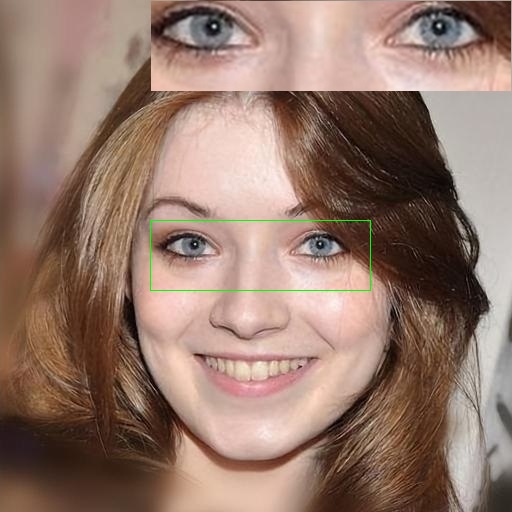}
    \\
    \includegraphics[width=0.142\linewidth]{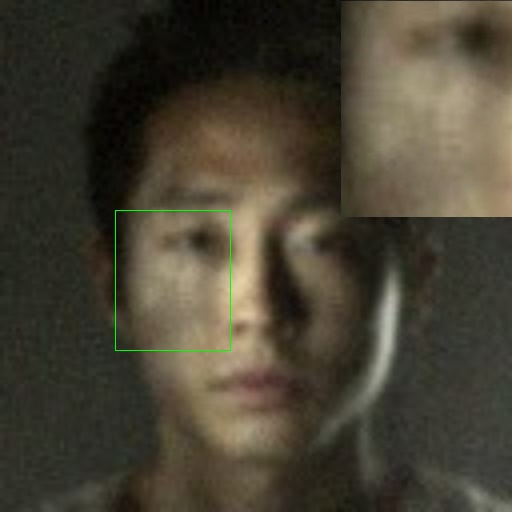}&
    \includegraphics[width=0.142\linewidth]{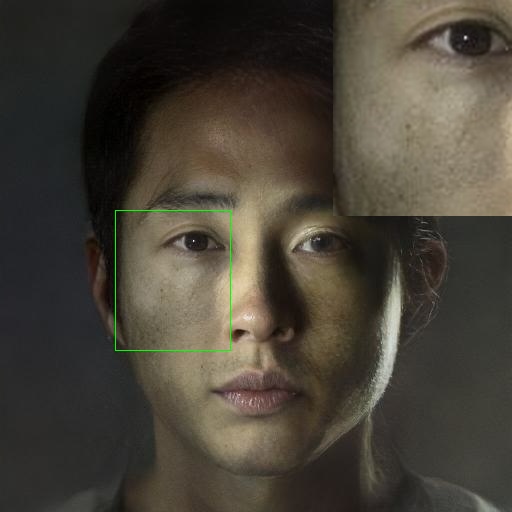}&
    \includegraphics[width=0.142\linewidth]{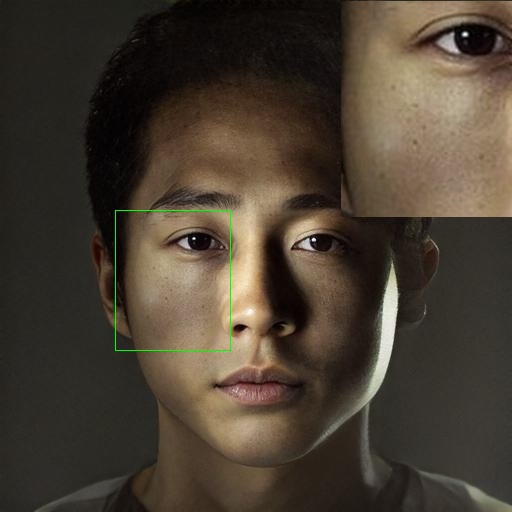}&
    \includegraphics[width=0.142\linewidth]{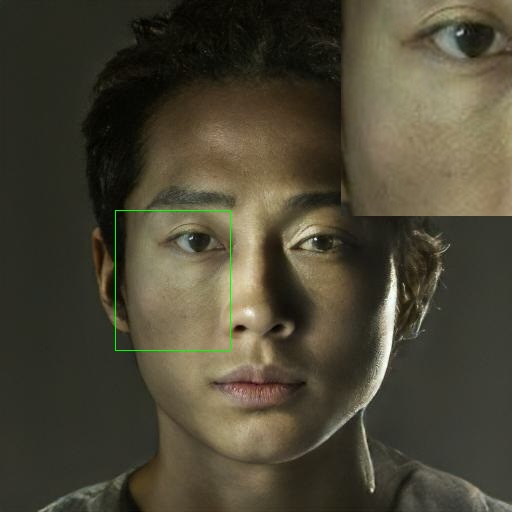}&
    \includegraphics[width=0.142\linewidth]{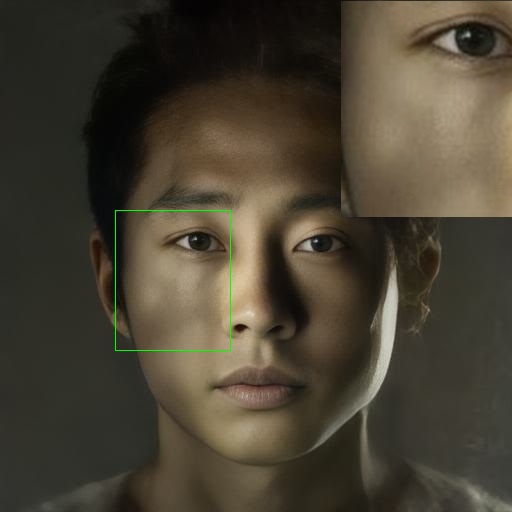}&
    \includegraphics[width=0.142\linewidth]{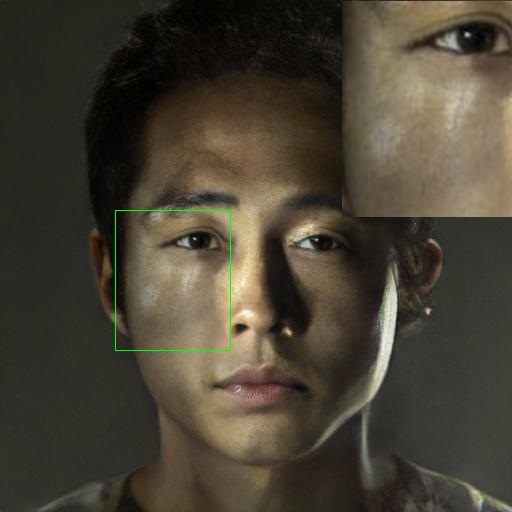}&
    \includegraphics[width=0.142\linewidth]{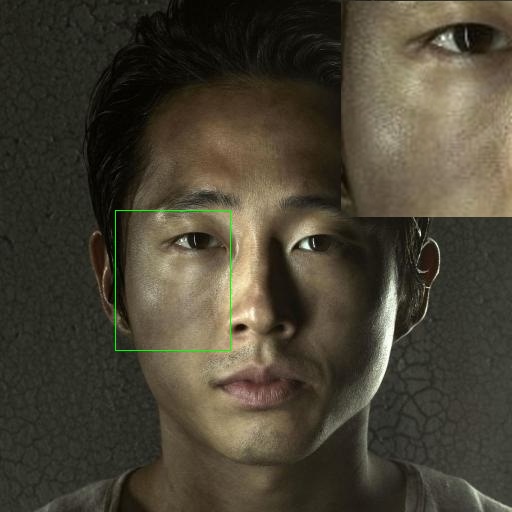}
    \\
    Input & DFDNet & GPEN & VQFR & GFP-GAN & \textbf{ENTED} & Ground Truth
  
    \end{tabular}
    \vspace{1mm}
    \caption{ The first row demonstrates a 8x blind face restoration. Our results display fewer distortions and align more closely with the original images, particularly in terms of the color of the pupils. The second row shows a 4x blind face restoration. Our approach reveals a skin texture that is more detailed and refined compared to what is achieved by current state-of-the-art methods.}
  \label{fig: comparison1}
  %\vspace{1mm}
\end{figure*}

\begin{equation}
\begin{aligned}
\hat{V}_{LQ} &= \phi_{hw}(\tilde{Z}_{LQ}, \tilde{Z}_{ref}, \tilde{Z}_{ref}), \\
\hat{V}_{ref} &= \phi_{hw}(\tilde{Z}_{ref}, \tilde{Z}_{LQ}, \tilde{Z}_{LQ}),  
\end{aligned}
\end{equation}
where $\hat{V}_{LQ}$, $\hat{V}_{ref}$ are the refined feature maps encoding exchanged information.

$\hat{V}_{LQ}$, $\hat{V}_{ref}$ are then reshaped to normal size with $c\times h\times w$ and re-scaled by learn-able scale factors $\gamma_{LQ} \in \mathbb{R}^{c}$ and $\gamma_{ref} \in \mathbb{R}^{c}$ respectively. To make the best use of high-quality information in the reference image, we concatenate both the input and reference's refined latent representations along the feature channel c to achieve a more informative and complete inference for the style code generation. 
The style code $\omega  \in \mathbb{R}^{c}$ is then obtained by embedding refined latent representations as
\begin{equation}
\omega = \Psi(concat[\gamma_{LQ}\hat{V}_{LQ} + \hat{Z}_{LQ}, \gamma_{ref}\hat{V}_{ref} + \hat{Z}_{ref}  ]) .
\end{equation}
where $\Psi(\cdot)$ is a linear layer.

\subsection{Loss Functions}
\label{loss_fn}
To train our model, we employ non-saturating adversarial loss \cite{karras2020analyzing} for restoring photo-realistic images, perceptual loss \cite{johnson2016perceptual} to evaluate the style difference, quantization loss \cite{esser2021taming} as discussed in Section \ref{VQD} and attention reconstruction loss \cite{ren2022neural} to supervise the texture extraction and distribution process.

\textbf{Adversarial loss} $\mathcal{L}_{adv}$.\cite{karras2020analyzing} To motivate the generator to prioritize entries in the natural image manifold and generate realistic textures, we train the generator with non-saturating adversarial loss. Denote the discriminator as $\mathcal{D}$ and the generator as $\mathcal{G}$. We use $\chi$ to denote the set of degraded images. Given a low-quality input image $I_{LQ}$ and a reference image $I_{ref}$, the adversarial loss is calculated as
\begin{equation}
\mathcal{L}_{adv} =  \mathbb{E}_{\substack{I_{LQ} \sim \chi\\}} log\biggl( 1 + exp\Bigl(-\mathcal{D}\bigl(\mathcal{G}(I_{LQ}, I_{ref})\bigr)\Bigr)\biggr).
\end{equation}

%\noindent where \lambda_(adv)\lambda_(adv) is the weighting factor for adversarial loss.

\textbf{Perceptual loss} $\mathcal{L}_{percep}$. \cite{johnson2016perceptual} Given a degraded image $I_{LQ}$ and a reference image $I_{ref}$, the restoration model $\mathbb{G}$ generates a restored image $I_{SR} = \mathbb{G}(I_{LQ}, I_{ref})$ . The restored image $I_{SR}$ is then compared with corresponding ground truth image $I_{gt}$ and evaluated by the $l_{1}$ difference between the pretrained VGG-19 \cite{simonyan2014very} activations:
\begin{equation}
\mathcal{L}_{percep} = \sum_{\substack{j\\}} \| \varphi_{j}(I_{SR}) - \varphi_{j}(I_{gt}) \|_{1},
\end{equation}
where $\varphi_{j}$ indicates the $j^{th}$ activation map of VGG-19.

\textbf{Attention reconstruction loss} $\mathcal{L}_{att}$. We use attention reconstruction loss \cite{ren2022neural} 
 to penalize inaccurate texture extracted from the reference image to facilitate the performance of the texture extraction and distribution process. As mentioned in Section \ref{sec3_1}, at each layer $l$ of the encoders, we can obtain an extraction matrix $\tilde{C}^l_e \in \mathbb{R}^{k\times HW/s^2}$ and a distribution matrix $\tilde{C}^l_d \in \mathbb{R}^{k\times HW/s^2}$, where $H$ and $W$ are the initial height and width of the input image and $s$ is the size downsample ratio at the $l$-th layer. Therefore the attention reconstruction loss is calculated as:
\begin{equation}
\mathcal{L}_{att} = \sum_{\substack{l\\}} \| I^{l  \downarrow}_{gt} -   I^{l\downarrow}_{ref}(\tilde{C}^l_e)^{\intercal} \tilde{C}^l_d\|_{1},
\end{equation}
where $I^{l  \downarrow_s}_{gt}$, $I^{l  \downarrow_s}_{ref}\in \mathbb{R}^{3\times HW/s^2}$ represent the bilinearly downsampled and dimension-compressed ground truth image and reference image corresponding to the $l^{th}$ layer, respectively.

\textbf{Total loss} $\mathcal{L}_{total}$. The overall loss is the sum of the losses mentioned above:
\begin{equation}
\mathcal{L}_{total} = \lambda_{adv}\mathcal{L}_{adv} + \lambda_{percep}\mathcal{L}_{percep} + \lambda_{q}\mathcal{L}_{q} + \lambda_{att}\mathcal{L}_{att},
\end{equation}
where  $\lambda_{adv}$, $\lambda_{percep}$, $\lambda_{q}$ and $\lambda_{att}$ are weighting factors corresponding to adversarial loss, perceptual loss, quantization loss, and attention reconstruction loss respectively. 

\section{Experiments}
\subsection{Experimental Setup}
\begin{table*}[h!]
    \centering
    \caption{Quantitative comparison with state-of-the-art blind face restoration methods on the modified CelebA-HQ test set. \textbf{\textcolor{red}{Red}} indicates the best performance while \textbf{\textcolor{blue}{Blue}} indicates the second best performance}
    \vspace{1mm}
    \renewcommand{\arraystretch}{1.2}
    \begin{tabular}{p{6mm}c@{\hspace{6mm}} *{7}{c@{\hspace{9mm}}}c}
    \hline
    &\textbf{Method} & \textbf{LPIPS} $\downarrow$ & \textbf{NIQE} $\downarrow$ &  \textbf{FID} $\downarrow$ & \textbf{Deg.} $\downarrow$  & \textbf{PSNR} $\uparrow$ & \textbf{SSIM} $\uparrow$  \\
    \hline
    &HiFaceGAN~\cite{yang2020hifacegan}   &0.3789 & 4.96 & 40.99 &35.13 & \textbf{\textcolor{blue}{25.57}}  & 0.6503  \\
    &PSFRGAN~\cite{chen2021progressive}   &0.2779 & 4.15 & 24.04 &36.12 & 25.11  & 0.6272  \\
    &DFDNet~\cite{li2020blind} &0.2910 &5.41 & 19.47 &30.48 &24.69   &0.6375  \\
    &GPEN~\cite{yang2021gan}       &0.2439  & 4.08 & 13.79 &\textbf{\textcolor{blue}{28.65}} & 25.11  & 0.6313     \\
    &VQFR~\cite{gu2022vqfr}         &0.2274 & \textbf{\textcolor{blue}{3.68}} & \textbf{\textcolor{red}{12.02}} &32.91 & 24.25  & 0.6371 \\
    &GFP-GAN~\cite{wang2021towards} &\textbf{\textcolor{blue}{0.2224}} & 4.23 & 13.63 &32.84 & 25.34  & \textbf{\textcolor{red}{0.6694}}     \\
    \hline
    &\textbf{ENTED} (Ours)       &\textbf{\textcolor{red}{0.2156}}  & \textbf{\textcolor{red}{3.60}}& \textbf{\textcolor{blue}{12.80}} & \textbf{\textcolor{red}{27.77}} & \textbf{\textcolor{red}{25.87}}   & \textbf{\textcolor{blue}{0.6664}}    \\
    \hline
    \end{tabular}
    \label{table: quantitative1}
\end{table*}
 We train our model from scratch on the FFHQ dataset \cite{karras2019style}, which contains 70,000 high-resolution images. As for the reference base blind face restoration, it is hard to find a corresponding reference image for each photo in a huge dataset. Therefore, we adopt~\cite{tov2021designing} to generate reference images during training by altering pose and age. All images are resized to 512 $\times$ 512 during training. Additionally, we employ the degradation model used in ~\cite{wang2021towards} to generate low-quality input images. The degradation model simulates low-quality images found in the actual world as a result of noises, defocus, jpeg compression, long-range sensing, and their combinations. 
As we propose a reference-based method for blind face restoration, which requires high-quality reference images, we use the CelebA-HQ test set and select the relevant image from real data distribution as the reference image for each individual test image in the dataset. There is no overlap between test data, including the reference images, and our training data. Table \ref{table: quantitative1} quantitatively demonstrates the 4x blind face restoration results conducted on this modified CelebA-HQ test set which consists of 2398 image pairs with corresponding reference images for quantitative and visual evaluation. As demonstrated in Table \ref{T:rwdata}, we also evaluate our model performance quantitatively on real-world data. For \textbf{\emph{LFW-Test}}, we carefully select 1,680 images with corresponding relevance from their own data distribution. For \textbf{\emph{CelebChild-Test}}, which consists of 180 child images, we take the adult image as the reference image for each corresponding child image. For \textbf{\emph{WebPhoto-Test}}, there are a total of 407 images in this dataset. As it is a dataset constructed by crawling images from the internet, it is difficult to find a corresponding reference image for each individual test data from the dataset. Therefore, we employ Tov et al. ~\cite{tov2021designing}  to generate reference images for each test image, the same as what we have done for the FFHQ training set.

\textbf{Implementation Setting}. The model generates images of 512 $\times$ 512 resolution with a given degraded input of size 512 $\times$ 512. The training batch size is set to 4. We augment the data by applying the degradation model. We employ the pretrained vector-quantized (VQ) dictionary from Gu et al.~\cite{gu2022vqfr}. The VQ dictionary contains 1024 entries with a code length of 256 for each latent code. The VQ dictionary is re-estimated via k-means clustering for every 2000 iterations. We also adopt 10,000 warm-up iterations before starting to adopt the VQ dictionary. Our model was trained using the Adam optimizer \cite{kingma2014adam} for 400,000 iterations, using a learning rate of $2 \times 10^{-3}$ for the generator and $1.882 \times 10^{-3}$ for the discriminator. The weighting factor $\lambda_{adv}$ , $\lambda_{percep}$, $\lambda_{q}$, $\lambda_{att}$ are set to 1.5, 1.0, 1.0 and 15.0 respectively. 

\begin{table*}[!h]
\caption{Quantitative Comparison with state-of-the-art blind face restoration methods on modified real-world LFW, CelebChild, WebPhoto dataset. \textbf{\textcolor{red}{Red}} indicates the best performance while \textbf{\textcolor{blue}{Blue}} indicates the second best performance }
\label{T:rwdata}
\begin{center}

\begin{tabular}{|c@{\hspace{10mm}}|c|c|c|c|c|c|c|}
\hline
% \multirow{1}{*}{\textbf{Dataset}}
% \multirow{2}{*}{\textbf{Models}}
 \textbf{Dataset} &\multicolumn{2}{c|}{\textbf{LFW-Test}}  & \multicolumn{2}{ c|}{\textbf{CelebChlid}} & \multicolumn{2}{ c|}{\textbf{WebPhoto}} \\
%\cmidrule{1-7}
\cline{1-7}
% \multirow{1}{*}{\textbf{Method}}
 \textbf{Method}
 &\textbf{FID} $\downarrow$
 & \textbf{NIQE} $\downarrow$
 &\textbf{FID} $\downarrow$
 & \textbf{NIQE} $\downarrow$
 &\textbf{FID} $\downarrow$
 & \textbf{NIQE} $\downarrow$
 \\
 \hline

  \hspace{5mm} HiFaceGAN~\cite{yang2020hifacegan}  &95.32 &4.94  &130.62  &4.99  &123.59 &5.48  \\ 
  \hspace{5mm} PSFRGAN~\cite{chen2021progressive} &70.57 &4.89  &114.95  &4.80  &90.19 &4.34  \\ 
  \hspace{5mm} DFDNet~\cite{li2020blind} &75.11 &5.81  &111.09  &5.81  &105.30 &5.70  \\ 
  \hspace{5mm} GPEN~\cite{yang2021gan} &70.74 &4.62  &110.70  &4.33  &95.60 &4.78  \\ 
  \hspace{5mm} VQFR~\cite{gu2022vqfr}  &\textbf{\textcolor{blue}{70.21}} &\textbf{\textcolor{blue}{3.98}}  &\textbf{\textcolor{blue}{106.18}}  &\textbf{\textcolor{red}{3.74}}  &\textbf{\textcolor{blue}{78.68}} &\textbf{\textcolor{red}{3.77}}  \\ 
  \hspace{5mm} GFP-GAN~\cite{wang2021towards} &\textbf{\textcolor{red}{69.04}} &4.77  &113.30 &4.55 &86.79 &4.40 \\ 
  \hline
  \hspace{5mm} \textbf{ENTED} (Ours)  & 77.76 & \textbf{\textcolor{red}{3.78}} & \textbf{\textcolor{red}{104.37}} &\textbf{\textcolor{blue}{3.79}} &\textbf{\textcolor{red}{76.24}}  &\textbf{\textcolor{blue}{3.91}} \\ \hline

\end{tabular}
\vspace{-4mm}
\end{center}
\end{table*}

\textbf{Evaluation Metrics}. We qualitatively evaluate our models based on the Peak Signal-to-Noise Ratio (PSNR), the Structural Similarity Index Measure (SSIM), the Learned Perceptual Image Patch Similarity (LPIPS) \cite{zhang2018unreasonable}, the Fréchet Inception Distance (FID) \cite{heusel2017gans}, the Natural Image Quality Evaluator (NIQE) \cite{mittal2012making} and the identity metric indicated as Deg., which calculates the face embedding angle between the restored face and the ground truth face using the embedding angle of ArcFace \cite{deng2019arcface}. For the LPIPS evaluation model, we adopt pretrained AlexNet \cite{krizhevsky2017imagenet} to evaluate the LPIPS value for all our experiments. 
\begin{table}[!h]
\caption{Ablation study results on Modified CelebA-HQ test set. \textit{Skip}: applying skip connection; \textit{Style}: applying modulated convolution; \textit{VQ}: applying vector-quantized dictionary; \textit{Refine}: applying latent space refinement. \textbf{\textcolor{red}{Red}} indicates the best performance while \textbf{\textcolor{blue}{Blue}} indicates the second best performance}
%\vspace{-1mm}
\label{T:ablation}
\centering
\hspace{-1mm}
\begin{tabular}{@{\hspace{0.5mm}}c@{\hspace{0.5mm}}|c@{\hspace{0.5mm}}|c@{\hspace{0.5mm}}|c@{\hspace{0.5mm}}|c@{\hspace{0.5mm}}|c@{\hspace{0.5mm}}|c@{\hspace{0.5mm}}|c@{\hspace{0.5mm}}}
\hline
\multirow{2}{*}{\textbf{\small Models}}
% \multirow{2}{*}{\textbf{Models}}
&\multicolumn{4}{c@{\hspace{0.3mm}}|}{\textbf{\small Configuration}}  & \multicolumn{3}{c@{\hspace{0.3mm}}}{\textbf{\small CelebA-Test}}\\ 
% \cmidrule{2-8}
\cline{2-8}

 &\textbf{\small Skip} 
 & \textbf{\small Style} 
 & \textbf{\small VQ} 
 & \textbf{\small Refine} 
 & \textbf{\small LPIPS} $\downarrow$
 & \textbf{\small FID} $\downarrow$
 & \textbf{\small NIQE} $\downarrow$ \\
 \hline

  Model 1&  & $\checkmark$  & $\checkmark$ &  &0.2762 &16.59 &3.64 \\ 
  Model 2& $\checkmark$  &  & $\checkmark$ &  &0.2185 &\textbf{\textcolor{red}{12.20}} & 3.73\\ 
  Model 3& $\checkmark$  & $\checkmark$ &  &  &\textbf{\textcolor{red}{0.2121}} &12.89 &3.81 \\ 
  Model 4& $\checkmark$ & $\checkmark$ & $\checkmark$ & &0.2178 &13.08 &\textbf{\textcolor{blue}{3.63}}\\ 
  \hline
  \small ENTED & $\checkmark$ & $\checkmark$ & $\checkmark$ & $\checkmark$ &\textbf{\textcolor{blue}{0.2156}}  &\textbf{\textcolor{blue}{12.80}} &\textbf{\textcolor{red}{3.60}}\\ \hline

\end{tabular}
\end{table}

\subsection{Comparison with Blind Face Restoration Baselines}
Our model is compared to several state-of-the-art blind face restoration baselines. We compare
our model with GFP-GAN \cite{wang2021towards}, VQFR \cite{gu2022vqfr}, GPEN \cite{yang2021gan}, DFDNet \cite{li2020blind}. PSFRGAN \cite{chen2021progressive} and HiFaceGAN \cite{yang2020hifacegan}. We adopt the official code and pre-trained weight for all the qualitative and quantitative comparisons. 

%, which are divided into two groups based on the nature of their input: reference-based super-resolution (RefSR), which requires input in two forms: degraded input and a high-quality reference image, and single image super-resolution (SISR), which requires only degraded input. For reference-based methods, we compare our model with C2-matching \cite{jiang2021robust}, MASA \cite{lu2021masa} and TTSR \cite{yang2020learning}. For single-image super-resolution techniques, there are two types: blind face restoration models that focus solely on recovering facial images and non-blind face restoration models that are broader and do not specify a particular aspect of renovation work. For blind restoration methods,

\subsubsection{Comparison on Modified CelebA-HQ Test}
The experimental results are the outcomes of 4x blind face restoration. As seen in Table \ref{table: quantitative1}, \textbf{ENTED} has the best LPIPS and NIQE. The improvement in LPIPS demonstrates that our method generates images that are perceptually nearer to ground truth than other cutting-edge methods. Furthermore, our model achieves the best NIQE, suggesting that our approach can generate face images with better high-quality details than other state-of-the-art approaches. The restored image is closer to the natural image distribution. Besides, our method preserves richer identity information from the given input, as indicated by the smallest face feature embedding angle degree. The best PSNR value in Table \ref{table: quantitative1} also suggests that our results have a closer pixel-wise value with the ground truth.

\subsubsection{Comparison on Real-world Datasets}
We conduct experiments on Modified \textbf{\emph{LFW-Test}}, \textbf{\emph{CelebChild-Test}} and \textbf{\emph{WebPhoto-Test}} separately.To compare our method to state-of-the-art methods in the extreme scenario for real-world data, we impose a bit of deterioration on the input data using the degradation model \cite{wang2021towards}. According to the quantitative findings in Table \ref{T:rwdata}, \textbf{ENTED} has relatively better FID results over three real-world datasets, indicating that our method can recover images near the true face distribution. On the other hand, \textbf{ENTED} outperforms the majority of other state-of-the-art methods across three real-world datasets in terms of NIQE.

\subsection{Ablation Study}

\subsubsection{Effect of Residual Connections}
The residual connection plays a crucial role in transmitting fidelity information from the content encoder to the decoder, ensuring the preservation of important identity details in the final restored image. By comparing Model 1 and Model 4 in Table \ref{T:ablation}, we observe that Model 1 performs better in terms of LPIPS and FID. This indicates that the output from the model without the residual connection is perceptually distant from the ground truth and true face distribution. The absence of the residual connection leads to a significant decrease in fidelity. However, by incorporating the residual connection into the decoder, the content encoder is able to provide complete fidelity information, resulting in facial details that are consistent with the ground truth distribution.

\subsubsection{Effect of Applying Modulated Convolution}
When comparing Model 2 and Model 4, we observe that Model 4 achieves lower LPIPS and NIQE values. This suggests that Model 4 is perceptually closer to the ground truth distribution and exhibits more high-quality details. The utilization of modulated convolution promotes the generation of high-frequency details, resulting in more photo-realistic images. However, in Model 4, there is no refinement of the latent space for degraded features, and the style code is directly generated from the degraded input's latent space. As a result, any defective features present in the input affect all modulated features, leading to inaccurate inference for fidelity details. This discrepancy contributes to a slightly higher FID in Model 4, indicating that the output of Model 4 deviates from the true face features.

\subsubsection{Effect of VQ Dictionary}
Since we are utilizing the high-quality details from the reference image, the effectiveness of the texture transfer process is vital for generating high-quality facial details in the restored image. The VQ dictionary provides high-quality semantic queries to help in texture distribution, resulting in effective texture transfer. When Model 3 and Model 4 are compared in Table \ref{T:ablation}, it is shown that the performance in NIQE improves when the VQ dictionary is included, indicating that the existence of VQ dictionary helps to produce more high-quality details.

\subsubsection{Effect of Latent Space Refinement}
The quality of the final restored image is closely related to the quality of the generated style code. As shown in Table \ref{T:ablation}, Model 4 does not employ the latent space refinement procedure and instead generates the style code straight from the latent space of degraded input. Erroneous information in the degraded input's latent space impacts the quality of the style code unless the input's latent space is rectified  using high-quality features from the reference prior. Refining the Latent space using reference prior aids in the generation of high-quality style code, resulting in photorealistic images with more high-quality details (e.g., best NIQE value, which favors high-quality details) and hence improved performance.

\section{Conclusion}
In this study, we have introduced a reference-based approach for blind face restoration. To address the blind face restoration problem, we have expanded and adapted the texture extraction and distribution framework proposed in controllable human image synthesis task. We have explored the network architecture and made modifications by incorporating a skip connection and utilizing modulated convolution. Additionally, we have integrated the VQ dictionary to facilitate the texture distribution process. To enhance the image restoration capabilities of the model, we have incorporated two types of attention mechanisms. A double attention mechanism has been employed for the texture transfer process, while the cross-attention technique has been utilized to refine the latent space and generate high-quality style codes.

%%%%%%%%% REFERENCES


\begin{thebibliography}{10}\itemsep=-1pt

\bibitem{chen2021progressive}
Chaofeng Chen, Xiaoming Li, Lingbo Yang, Xianhui Lin, Lei Zhang, and Kwan-Yee~K Wong.
\newblock Progressive semantic-aware style transformation for blind face restoration.
\newblock In {\em CVPR}, pages 11896--11905, 2021.

\bibitem{chen2022blind}
Chaofeng Chen, Xinyu Shi, Yipeng Qin, Xiaoming Li, Xiaoguang Han, Tao Yang, and Shihui Guo.
\newblock Blind image super resolution with semantic-aware quantized texture prior.
\newblock {\em arXiv preprint arXiv:2202.13142}, 2022.

\bibitem{chen2018fsrnet}
Yu Chen, Ying Tai, Xiaoming Liu, Chunhua Shen, and Jian Yang.
\newblock Fsrnet: End-to-end learning face super-resolution with facial priors.
\newblock In {\em CVPR}, pages 2492--2501, 2018.

\bibitem{chu2022nafssr}
Xiaojie Chu, Liangyu Chen, and Wenqing Yu.
\newblock Nafssr: Stereo image super-resolution using nafnet.
\newblock In {\em CVPR}, pages 1239--1248, 2022.

\bibitem{deng2019arcface}
Jiankang Deng, Jia Guo, Niannan Xue, and Stefanos Zafeiriou.
\newblock Arcface: Additive angular margin loss for deep face recognition.
\newblock In {\em CVPR}, pages 4690--4699, 2019.

\bibitem{dogan2019exemplar}
Berk Dogan, Shuhang Gu, and Radu Timofte.
\newblock Exemplar guided face image super-resolution without facial landmarks.
\newblock In {\em CVPRW}, pages 0--0, 2019.

\bibitem{esser2021taming}
Patrick Esser, Robin Rombach, and Bjorn Ommer.
\newblock Taming transformers for high-resolution image synthesis.
\newblock In {\em CVPR}, pages 12873--12883, 2021.

\bibitem{gu2020image}
Jinjin Gu, Yujun Shen, and Bolei Zhou.
\newblock Image processing using multi-code gan prior.
\newblock In {\em CVPR}, pages 3012--3021, 2020.

\bibitem{gu2022vqfr}
Yuchao Gu, Xintao Wang, Liangbin Xie, Chao Dong, Gen Li, Ying Shan, and Ming-Ming Cheng.
\newblock Vqfr: Blind face restoration with vector-quantized dictionary and parallel decoder.
\newblock {\em arXiv preprint arXiv:2205.06803}, 2022.

\bibitem{heusel2017gans}
Martin Heusel, Hubert Ramsauer, Thomas Unterthiner, Bernhard Nessler, and Sepp Hochreiter.
\newblock Gans trained by a two time-scale update rule converge to a local nash equilibrium.
\newblock {\em NIPS}, 30, 2017.

\bibitem{jiang2021robust}
Yuming Jiang, Kelvin~CK Chan, Xintao Wang, Chen~Change Loy, and Ziwei Liu.
\newblock Robust reference-based super-resolution via c2-matching.
\newblock In {\em CVPR}, pages 2103--2112, 2021.

\bibitem{johnson2016perceptual}
Justin Johnson, Alexandre Alahi, and Li Fei-Fei.
\newblock Perceptual losses for real-time style transfer and super-resolution.
\newblock In {\em ECCV}, pages 694--711. Springer, 2016.

\bibitem{karras2019style}
Tero Karras, Samuli Laine, and Timo Aila.
\newblock A style-based generator architecture for generative adversarial networks.
\newblock In {\em CVPR}, pages 4401--4410, 2019.

\bibitem{karras2020analyzing}
Tero Karras, Samuli Laine, Miika Aittala, Janne Hellsten, Jaakko Lehtinen, and Timo Aila.
\newblock Analyzing and improving the image quality of stylegan.
\newblock In {\em CVPR}, pages 8110--8119, 2020.

\bibitem{kim2019progressive}
Deokyun Kim, Minseon Kim, Gihyun Kwon, and Dae-Shik Kim.
\newblock Progressive face super-resolution via attention to facial landmark.
\newblock {\em arXiv preprint arXiv:1908.08239}, 2019.

\bibitem{kingma2014adam}
Diederik~P Kingma and Jimmy Ba.
\newblock Adam: A method for stochastic optimization.
\newblock {\em arXiv preprint arXiv:1412.6980}, 2014.

\bibitem{krizhevsky2017imagenet}
Alex Krizhevsky, Ilya Sutskever, and Geoffrey~E Hinton.
\newblock Imagenet classification with deep convolutional neural networks.
\newblock {\em Communications of the ACM}, 60(6):84--90, 2017.

\bibitem{lancucki2020robust}
Adrian {\L}a{\'n}cucki, Jan Chorowski, Guillaume Sanchez, Ricard Marxer, Nanxin Chen, Hans~JGA Dolfing, Sameer Khurana, Tanel Alum{\"a}e, and Antoine Laurent.
\newblock Robust training of vector quantized bottleneck models.
\newblock In {\em IJCNN}, pages 1--7. IEEE, 2020.

\bibitem{li2020blind}
Xiaoming Li, Chaofeng Chen, Shangchen Zhou, Xianhui Lin, Wangmeng Zuo, and Lei Zhang.
\newblock Blind face restoration via deep multi-scale component dictionaries.
\newblock In {\em ECCV}, pages 399--415. Springer, 2020.

\bibitem{EBFR_Multi_Exemplar}
Xiaoming Li, Wenyu Li, Dongwei Ren, Hongzhi Zhang, Meng Wang, and Wangmeng Zuo.
\newblock Enhanced blind face restoration with multi-exemplar images and adaptive spatial feature fusion.
\newblock In {\em CVPR}, June 2020.

\bibitem{Learning_Warped}
Xiaoming Li, Ming Liu, Yuting Ye, Wangmeng Zuo, Liang Lin, and Ruigang Yang.
\newblock Learning warped guidance for blind face restoration.
\newblock In {\em ECCV}, September 2018.

\bibitem{Dual_Memory}
Xiaoming Li, Shiguang Zhang, Shangchen Zhou, Lei Zhang, and Wangmeng Zuo.
\newblock Learning dual memory dictionaries for blind face restoration.
\newblock {\em TPAMI}, 2022.

\bibitem{lu2021masa}
Liying Lu, Wenbo Li, Xin Tao, Jiangbo Lu, and Jiaya Jia.
\newblock Masa-sr: Matching acceleration and spatial adaptation for reference-based image super-resolution.
\newblock In {\em CVPR}, pages 6368--6377, 2021.

\bibitem{menon2020pulse}
Sachit Menon, Alexandru Damian, Shijia Hu, Nikhil Ravi, and Cynthia Rudin.
\newblock Pulse: Self-supervised photo upsampling via latent space exploration of generative models.
\newblock In {\em CVPR}, pages 2437--2445, 2020.

\bibitem{mittal2012making}
Anish Mittal, Rajiv Soundararajan, and Alan~C Bovik.
\newblock Making a “completely blind” image quality analyzer.
\newblock {\em IEEE Signal processing letters}, 20(3):209--212, 2012.

\bibitem{ren2022neural}
Yurui Ren, Xiaoqing Fan, Ge Li, Shan Liu, and Thomas~H Li.
\newblock Neural texture extraction and distribution for controllable person image synthesis.
\newblock In {\em CVPR}, pages 13535--13544, 2022.

\bibitem{richardson2021encoding}
Elad Richardson, Yuval Alaluf, Or Patashnik, Yotam Nitzan, Yaniv Azar, Stav Shapiro, and Daniel Cohen-Or.
\newblock Encoding in style: a stylegan encoder for image-to-image translation.
\newblock In {\em CVPR}, pages 2287--2296, 2021.

\bibitem{shen2018deep}
Ziyi Shen, Wei-Sheng Lai, Tingfa Xu, Jan Kautz, and Ming-Hsuan Yang.
\newblock Deep semantic face deblurring.
\newblock In {\em CVPR}, pages 8260--8269, 2018.

\bibitem{simonyan2014very}
Karen Simonyan and Andrew Zisserman.
\newblock Very deep convolutional networks for large-scale image recognition.
\newblock {\em arXiv preprint arXiv:1409.1556}, 2014.

\bibitem{tov2021designing}
Omer Tov, Yuval Alaluf, Yotam Nitzan, Or Patashnik, and Daniel Cohen-Or.
\newblock Designing an encoder for stylegan image manipulation.
\newblock {\em ACM Transactions on Graphics (TOG)}, 40(4):1--14, 2021.

\bibitem{vaswani2017attention}
Ashish Vaswani, Noam Shazeer, Niki Parmar, Jakob Uszkoreit, Llion Jones, Aidan~N Gomez, {\L}ukasz Kaiser, and Illia Polosukhin.
\newblock Attention is all you need.
\newblock {\em Advances in neural information processing systems}, 30, 2017.

\bibitem{wan2020bringing}
Ziyu Wan, Bo Zhang, Dongdong Chen, Pan Zhang, Dong Chen, Jing Liao, and Fang Wen.
\newblock Bringing old photos back to life.
\newblock In {\em CVPR}, pages 2747--2757, 2020.

\bibitem{wang2021towards}
Xintao Wang, Yu Li, Honglun Zhang, and Ying Shan.
\newblock Towards real-world blind face restoration with generative facial prior.
\newblock In {\em CVPR}, pages 9168--9178, 2021.

\bibitem{wang2022restoreformer}
Zhouxia Wang, Jiawei Zhang, Runjian Chen, Wenping Wang, and Ping Luo.
\newblock Restoreformer: High-quality blind face restoration from undegraded key-value pairs.
\newblock In {\em CVPR}, pages 17512--17521, 2022.

\bibitem{xie2020feature}
Yanchun Xie, Jimin Xiao, Mingjie Sun, Chao Yao, and Kaizhu Huang.
\newblock Feature representation matters: End-to-end learning for reference-based image super-resolution.
\newblock In {\em ECCV}, pages 230--245. Springer, 2020.

\bibitem{yang2020learning}
Fuzhi Yang, Huan Yang, Jianlong Fu, Hongtao Lu, and Baining Guo.
\newblock Learning texture transformer network for image super-resolution.
\newblock In {\em CVPR}, pages 5791--5800, 2020.

\bibitem{yang2020hifacegan}
Lingbo Yang, Shanshe Wang, Siwei Ma, Wen Gao, Chang Liu, Pan Wang, and Peiran Ren.
\newblock Hifacegan: Face renovation via collaborative suppression and replenishment.
\newblock In {\em Proceedings of the 28th ACM International Conference on Multimedia}, pages 1551--1560, 2020.

\bibitem{yang2021gan}
Tao Yang, Peiran Ren, Xuansong Xie, and Lei Zhang.
\newblock Gan prior embedded network for blind face restoration in the wild.
\newblock In {\em CVPR}, pages 672--681, 2021.

\bibitem{yu2018face}
Xin Yu, Basura Fernando, Bernard Ghanem, Fatih Porikli, and Richard Hartley.
\newblock Face super-resolution guided by facial component heatmaps.
\newblock In {\em ECCV}, pages 217--233, 2018.

\bibitem{zhang2018unreasonable}
Richard Zhang, Phillip Isola, Alexei~A Efros, Eli Shechtman, and Oliver Wang.
\newblock The unreasonable effectiveness of deep features as a perceptual metric.
\newblock In {\em CVPR}, pages 586--595, 2018.

\bibitem{zhang2019image}
Zhifei Zhang, Zhaowen Wang, Zhe Lin, and Hairong Qi.
\newblock Image super-resolution by neural texture transfer.
\newblock In {\em CVPR}, pages 7982--7991, 2019.

\bibitem{zhao2022rethinking}
Yang Zhao, Yu-Chuan Su, Chun-Te Chu, Yandong Li, Marius Renn, Yukun Zhu, Changyou Chen, and Xuhui Jia.
\newblock Rethinking deep face restoration.
\newblock In {\em CVPR}, pages 7652--7661, 2022.

\bibitem{zhou2022towards}
Shangchen Zhou, Kelvin~CK Chan, Chongyi Li, and Chen~Change Loy.
\newblock Towards robust blind face restoration with codebook lookup transformer.
\newblock {\em arXiv preprint arXiv:2206.11253}, 2022.

\end{thebibliography}


{\small

}
\end{document}